# High-Level Plan for Behavioral Robot Navigation with Natural Language Directions and R-NET

Amar Shrestha[*], Krittaphat Pugdeethosapol[*], Haowen Fang, Qinru Qiu

Department of Electrical Engineering & Computer Science, Syracuse University, USA
{amshrest, kpugdeet, hfang02, qiqiu}@syr.edu

## Abstract

When the navigational environment is known, it can be represented as a graph where landmarks are nodes, the robot behaviors that move from node to node are edges, and the route is a set of behavioral instructions. The route path from source to destination can be viewed as a class of combinatorial optimization problems where the path is a sequential subset from a set of discrete items. The pointer network is an attention-based recurrent network that is suitable for such a task. In this paper, we utilize a modified R-NET with gated attention and self-matching attention translating natural language instructions to a high-level plan for behavioral robot navigation by developing an understanding of the behavioral navigational graph to enable the pointer network to produce a sequence of behaviors representing the path. Tests on the navigation graph dataset show that our model outperforms the state-of-the-art approach for both known and unknown environments.

## Introduction

In a complex but known environment, an automated robot needs to follow a route to reach the destination from a starting point. When the environment is described in terms of a graph (Sepulveda et al. 2018) where landmarks are nodes and the robot behaviors that moves it from one node to its adjacent nodes are edges, the route is a set of behavioral instructions (go left, go right, go forward, etc.). The route path from source to destination can be viewed as a class of combinatorial optimization problems where the path is a sequential subset from a set of discrete items.

Thus, this robot navigation application should take the environment or behavioral graph and the source-destination pair along with natural language directions as input and produce a sequence of behavioral instructions to reach the destination from the source. In terms of Neural Program Learning, this can be taken as an example-driven induction (Devlin et al. 2017).

Sequence-to-sequence paradigm of recurrent neural networks (RNNs) have been a staple of neural network architectures for learning functions over sequences (encoder) from examples and producing sequences of outputs (decoder). The content-based attentional mechanism has also been used to provide contextual information from the encoded input to the decoder. But these methods require a fixed-sized output dictionary which is not suitable for combinatorial problems where the size of the output dictionary depends on the length of the input sentence. This problem can be resolved by utilizing Pointer Networks (Vinyals et al. 2015) which augment the attention mechanism to create pointers to the input elements. The augmentation is simple: instead of blending the hidden encoder units using attention to provide context at each decoder step, utilize the attention as a pointer to select a member of the input sequence as the output.

In this work, we utilize the behavioral graph dataset from (Zang et al. 2018). Given the complexity of the problem, we move from the straight-forward sequence-to-sequence architecture in the original pointer network in (Vinyals et al. 2015) to a modification of R-Net (Wang et al. 2017) with gated-attention and self-matching attention mechanisms to:

1. Incorporate an understanding of the behavioral graph with the source-destination pair and natural language directions.
2. Aggregate evidence from the entire graph to infer the route from source to destination based on natural language directions.

## Related work

In order for an automated robot to follow a route based on navigation commands to reach the destination, the robot needs to be able to process and understand the natural language direction, as well as translating into plans or commands to execute. The most common approaches can be divided into three categories, manually parsing commands, constraining language descriptions, and statistical machine translation methods (Zang et al. 2018).

---

[*] These authors contributed equally to this work.

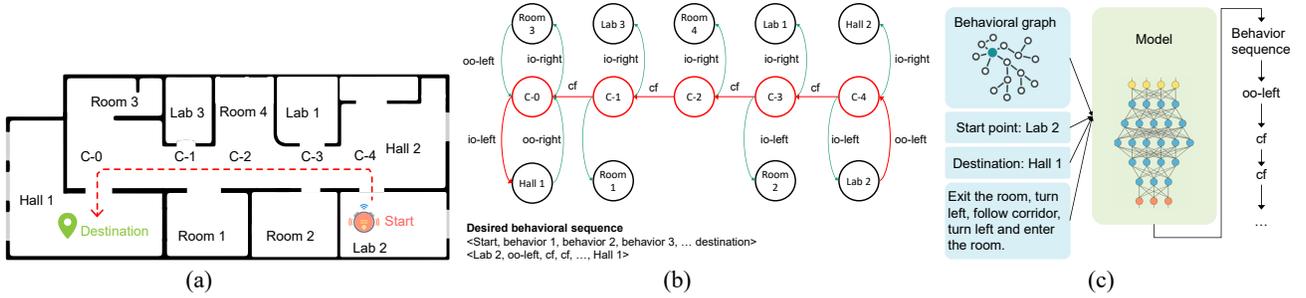

Figure 1. (a) Map of an environment. (b) Its behavioral navigation graph and desired behavioral sequence. (c) Problem setting. The red part of (b) corresponds to the representation of the route highlighted in red in (a). The codes "oo-left", "cf", "cf", and "io-left", correspond to the behaviors "Exit the room, turn left", "follow the corridor", "follow the corridor", "turn left and enter the room" respectively.

Manually parsing commands is a straight forward method to translate natural language into command, but it is inapplicable in a real-world situation. For example, (Levit and Roy 2007) proposed a method to translate the spatial language to a map navigational task by manually parsing natural language instructions into navigation information units (positions, orientations, moves, turns, and location references) sequence for navigation.

In constraining language descriptions, the space of input descriptions is limited to aid the translation into execution commands. (Schulz et al. 2015) proposed a symbolic navigation system that utilizes symbolic information extracted from door label and map environments to navigate robot from source to destination.

The statistical machine translation is the most recent approach translating the natural language to robot navigation instructions by utilizing translation rules created from a corpus of data. For example, (Matuszek et al. 2010) parse natural language instructions into a grammar liked language sequence by using the word alignment-based semantic parser (Wong and Mooney 2007).

In this work, we mainly focus on end-to-end learning to translate free-from natural language instruction to a high-level plan for behavioral robot navigation by utilizing a sequence to sequence neural network for natural language processing.

## Experimental Task

### Behavioral Graph

The navigation environment in the dataset (Zang et al. 2018) consists of 7 types of semantic locations, 12 types of behaviors shown in Table 1., and 20 different types of landmarks. A location in the environment can be a room, a lab, an office, a kitchen, a hall, a corridor, or a bathroom. As we require unique sets of elements in the graph and as the behaviors are not unique to its location, we encode the environment into a behavioral graph $m$ into unique triplets $T_l = <p_i; b_l; p_j>$, where $p_i$ and $p_j$ are adjacent nodes in the graph, and the edge $b_l$ is an executable behavior to navigate from $p_i$ to $p_j$. In this work, as opposed to (Zang et al. 2018), we remove the landmarks as they do not provide any more information and increases the dimensionality of the behavioral graph.

| Behavior | Description |
|---|---|
| oo<d> | Go out of the current place and turn <d> |
| io<d> | Turn <d> and enter the place straight ahead |
| oio | Exit current place and enter straight ahead |
| <d>t | Turn <d> at the intersection |
| cf | Follow (or go straight down) the corridor |
| sp | Go straight at a T intersection |
| st | Go straight through the corridor |
| ch<d> | Cross the hall and turn <d> |

Table 1: Behaviors (edges) of the navigation graphs considered in this work. The direction <d> can be left or right (Zang et al. 2018).

An example from the dataset and problem sets are shown in Figure 1. (a) shows the environment with the source and destination represented by red and green symbol respectively, and the red line showing the desired path between them which correspond to natural language direction. (b) shows the result behavioral navigation graph and the desired behavior sequence. (c) shows the problem setting where the inputs of the model are a behavioral graph, source-destination pair, and natural language instruction for the target path such as "*Exit the room, turn left, follow the corridor, turn left and enter the room*". The dataset consists of 8066 pairs of navigation plans and source-destination for training. This training data was collected from 88 unique simulated environments, totaling 6064

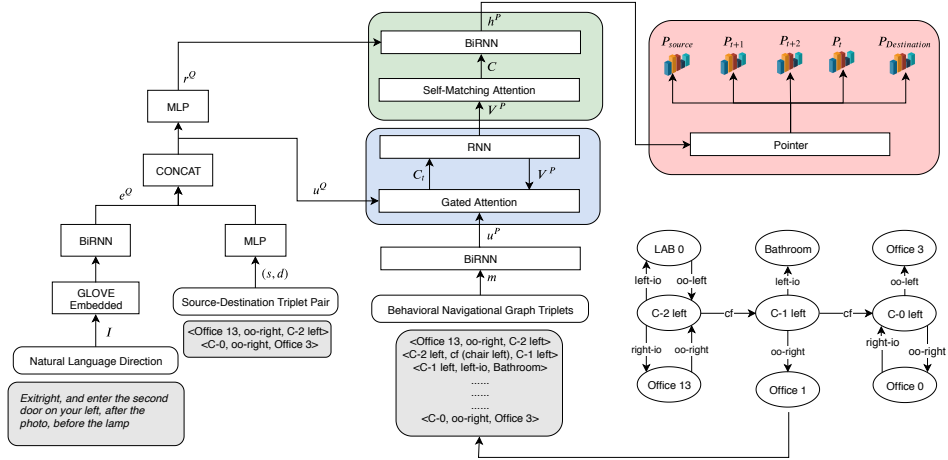

Figure 2. Modified R-NET Structure with natural language direction.

distinct navigation plans. The dataset consists of two test sets

(1) Test-repeated: This test set contains 1012 pairs of navigation plans and source-destination pair. These routes are not part of the training set; however, they are collected using environments that are part of the training set.

(2) Test-new: This test set contains 946 samples collected using environments that are not part of the training set. In the training set, the largest set of triplets is 500 and the smallest set is 200. Thus, we limit the training set to 300 maximum triplets. The graph with fewer triplets is padded and the graph with more than 300 triplets is cut but ensuring that the target sequence of triplets is included in the 300.

## Problem Formulation

The task in the work is to build a model to extrapolate the path from the source to the destination with natural language direction in a behavioral navigational graph for an indoor environment. We provide the model, source-destination triplet pair $(s, d)$, natural language direction $I$, and behavioral navigational graph $m$. Formally, we construct a model to predict the correct sequence of triplets to provide the correct sequence of behaviors $(b_1, b_2, ...)$ based on the previously unseen input of $(m, s, d, I)$. From a supervised learning perspective, the goal is then to estimate:

$$\underset{T_1,...,T_t}{argmax} \, P(T_1, ... T_t | m, s, d, I) \quad (1)$$

From the dataset, the input-output pair is $\{x_i, y_i) \mid 0 \leq i \leq N\}$ where $x_i = (m, s, d, I)_i$ and $y_i = (b_1, ..., b_t)_i \in (T_1, ..., T_t)_i$.

## Proposed Work

Figure 2. shows the overview of the modified R-NET architecture (Wang et al. 2017) that we utilize in this work. First, the behavioral navigational graph $m$ is processed by a bidirectional recurrent network (BiRNN) (Schuster and Paliwal 1997) to get $u^P$. The source-destination triplet pair $(s, d)$ is separately processed by a multilayer perceptron (MLP) with ReLU activation. The natural language direction $I$ is separated into words and embedded using GLOVE embedded vectors (Pennington et al. 2014), then processed by BiRNN to get $e^Q$. Second, we concatenate the processed source-destination triple pair with processed natural language direction to get $u^Q$:

$$u^Q = concat(MLP_Q(s, d), e^Q) \quad (2)$$

Third, we pass $u^P$ and $u^Q$ into a gated attention-based recurrent network to incorporate natural language direction and source-destination information into a behavioral navigational graph representation. Then, we refine the graph representation by aggregating the evidence from the whole navigational graph utilizing self-matching attention. This is then fed into the decoder layer with a pointer network containing Gated Recurrent Units (GRU) to produce predictions over the input triplets in the behavioral navigational graph.

The main differences from the original R-NET architecture are in the encoder where we use MLP and BiRNN to produce the new representation $u^Q$ and the pointer network in the decoder where the initial hidden state $(r^Q)$ of the BiRNN for the pointer network is determined from the encoding of the encode $u^Q$:

$$r^Q = MLP(u^Q) \tag{3}$$

The other parts remaining the same as the original R-NET paper and we use the same notations to ensure that the differences are visible.

## Evaluation

### Training Details

The training procedure used in this is straight-forward. We utilize the entire training set containing 8066 samples for training and the entire test-repeated set containing 1012 samples for validation. The summary of the dataset is shown in Table 2.

| Dataset | # Single | #Double | Total |
|---|---|---|---|
| Training | 4062 | 2002 | 8066 |
| Test-Repeated | 944 | 34 | 1012 |
| Test-New | 962 | 0 | 962 |

Table 2: Statistic of behavioral navigation dataset (Zang et al. 2018). "# Single" indicate the plans with single instruction. "# Double" indicate the plans with double instructions.

We utilize backpropagation with the ADAM optimizer to optimize the network. We also utilize a variational dropout as used in the original R-NET architecture. The evaluations were performed in both test-repeated and test-new set. The model is trained for 45 epochs. The number of units in BiRNN is set to 100 and there are 3 GRU layers in the BiRNN.

### Evaluation metrics

We utilize four metrics to evaluate the model as in (Zang et al. 2018):
- Exact Match (EM): If the predicted plan exactly matches the ground truth, the value is 1 and it is 0 otherwise.
- F1 score (F1): The harmonic mean of precision and recall.
- Edit Distance (ED): The minimum number of changes need to be made in order to transform a predicted sequence to ground truth sequence.
- Goal Match (GM): If a predicted sequence reaches the ground truth destination, the value is 1 even it has a different sequence and it is 0 otherwise.

### Results

As shown in Table 3, our model outperforms the original work (Zang et al. 2018) in both test-repeated and test-new set. This is because of modified R-NET, gated-attention and self-matching attention that produce a better understanding of the navigational graph, and pointer network to produce the output behavior sequence by producing a probability over the input graph instead of producing probabilities over the list of behaviors.

We can also see that our model performance does not drop when applying to test-new set compare to (Zang et al. 2018) in which EM and GM significantly drop.

| Model | Test-Repeated Set | | | |
|---|---|---|---|---|
| | EM ↑ | F1 ↑ | ED ↓ | GM ↑ |
| Zang et al. | 61.71 | 93.54 | 0.75 | 61.36 |
| Our | **72.65** | **95.56** | **0.31** | **85.15** |

| Model | Test-New Set | | | |
|---|---|---|---|---|
| | EM ↑ | F1 ↑ | ED ↓ | GM ↑ |
| Zang et al. | 41.71 | 90.22 | 1.22 | 41.81 |
| Our | **75.00** | **94.61** | **0.42** | **89.06** |

Table 3: Performance of our models compare to in (Zang et al. 2018) on the test datasets. EM and GM report percentages, and ED corresponds to average edit distance. The symbol ↑ indicates that higher results are better in the corresponding column; ↓ indicates that lower is better.

## Conclusion

In this paper, we utilize the behavioral graph dataset from (Zang et al. 2018) and given the complexity of the problem, we move from the straight-forward sequence-to-sequence architecture in the original pointer network to a modification of R-Net with gated attention and self-matching attention mechanisms. We utilize the graph, natural language direction, and the source-destination triplet pair as inputs to produce a sequence of behaviors.